\documentclass{svproc}
\usepackage{url}
\usepackage{booktabs}
\usepackage{amsmath}
\usepackage{subcaption}
\usepackage{graphicx}
\usepackage{listings}
\usepackage{pdflscape}
\usepackage{textcomp}
\usepackage{multirow}
\usepackage{threeparttable}
\usepackage{tabularx}

\begin{document}
\mainmatter            

\title{Solar Radiation Prediction in the UTEQ based on Machine Learning Models}

\titlerunning{Prediction of solar radiation in the UTEQ} 
\author{Jordy Anchundia Troncoso\inst{1} \and Ángel Torres Quijije\inst{1}
Byron Oviedo\inst{2} \and Cristian Zambrano-Vega\inst{1}\thanks{Corresponding author: czambrano@uteq.edu.ec}}
\authorrunning{Zambrano-Vega et al.}

\tocauthor{Jordy Anchundia Troncoso, Ángel Torres Quijije, Byron Oviedo, Cristian Zambrano-Vega}

\institute{State Technical University of Quevedo, Los Ríos, Ecuador\\
	Department of Engineering Science\\
	\email{\{jordy.anchundia2018, atorres, czambrano\}@uteq.edu.ec}\\
	\and
	State Technical University of Quevedo, Los Ríos, Ecuador\\
	Department of Graduate Programs\\
	\email{boviedo@uteq.edu.ec}
}
\maketitle              

\begin{abstract}

This research explores the effectiveness of various Machine Learning (ML) models used to predicting solar radiation at the Central Campus of the State Technical University of Quevedo (UTEQ). The data was obtained from a pyranometer, strategically located in a high area of the campus. This instrument continuously recorded solar irradiance data since 2020, offering a comprehensive dataset encompassing various weather conditions and temporal variations. After a correlation analysis, temperature and the time of day were identified as the relevant meteorological variables that influenced the solar irradiance. Different machine learning algorithms such as Linear Regression, K-Nearest Neighbors, Decision Tree, and Gradient Boosting were compared using the evaluation metrics Mean Squared Error (MSE), Root Mean Squared Error (RMSE), Mean Absolute Error (MAE), and the Coefficient of Determination ($R^2$). The study revealed that Gradient Boosting Regressor exhibited superior performance, closely followed by the Random Forest Regressor. These models effectively captured the non-linear patterns in solar radiation, as evidenced by their low MSE and high R² values. With the aim of assess the performance of our ML models, we developed a web-based tool for the Solar Radiation Forecasting in the UTEQ available at \url{http://https://solarradiationforecastinguteq.streamlit.app/}. The  results obtained demonstrate the effectiveness of our ML models in solar radiation prediction and contribute a practical utility in real-time solar radiation forecasting, aiding in efficient solar energy management.

\keywords{Solar Radiation, Machine Learning, Forecasting, UTEQ}
\end{abstract}

\section{Introduction}

Solar radiation, a critical source of energy for the Earth, profoundly influences climate, weather patterns, and renewable energy production. Accurate prediction of solar radiation is essential for optimizing the efficiency of solar energy systems and managing energy resources effectively. 

Recent advancements in Machine Learning (ML) have revolutionized the field of meteorology and renewable energy. ML's ability to handle large volumes of data and its prowess in identifying complex patterns make it an ideal tool for predicting solar radiation. This article explores the application of various ML models to predict solar radiation at the Central Campus of the State Technical University of Quevedo (UTEQ in Spanish) \url{https://www.uteq.edu.ec/}, highlighting the potential of these technologies in real-world scenarios.

This study is significant as it not only contributes to the body of knowledge in solar radiation prediction but also demonstrates the practical application of ML in renewable energy management. The findings of this research could be instrumental in improving solar energy utilization, thereby supporting sustainable energy initiatives.

\section{Materials and Methods}

\subsection{Data Collection}
\begin{figure}[!ht]
	\includegraphics[width=4cm]{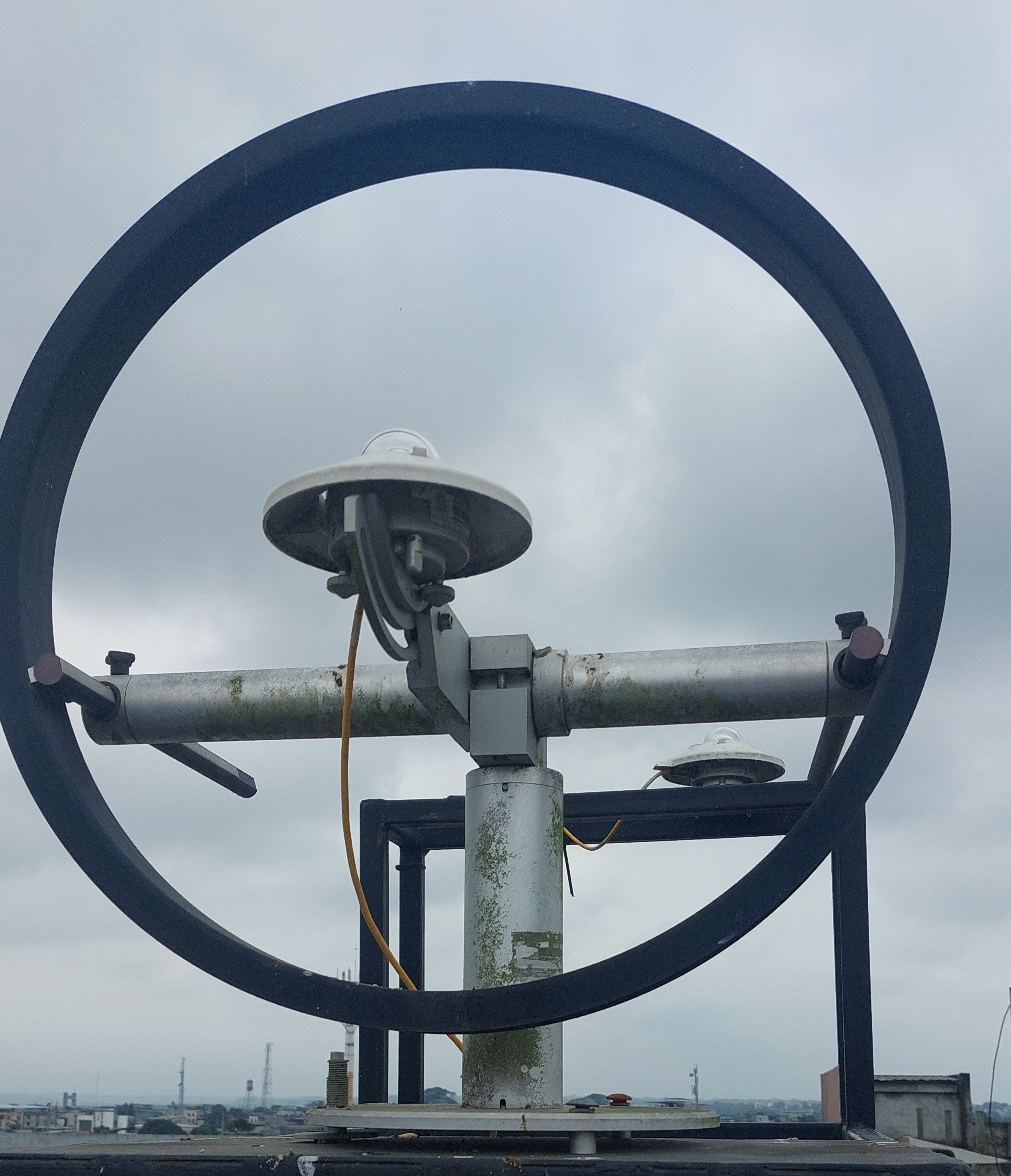}
	\centering
	\caption{Pyranometer installed in one of the Academic Buildings of the UTEQ}
	\label{fig:pyranometer}
\end{figure}

The data collection process was pivotal in developing an accurate predictive model for solar radiation. Primary data was obtained from a pyranometer, strategically located at the Universidad Técnica Estatal de Quevedo (UTEQ) (Fig. \ref{fig:pyranometer}).

This instrument continuously recorded solar irradiance data since 2020, offering a comprehensive dataset encompassing various weather conditions and temporal variations. The data, sampled at 5 minutes intervals, included key parameters like solar radiation intensity and ambient temperature. To ensure data integrity, regular maintenance checks were conducted on the pyranometer, and anomalous readings were flagged for review. 

\subsection{Data Description}

The dataset primarily consisted of data measured by a pyranometer every 5 minutes. The key fields included:

\begin{itemize}
	\item \textbf{Solar Irradiance}: Quantitative, continuous data representing the solar radiation intensity, it is recorded in watts per square metre (W/m²) units.
	\item \textbf{Ambient Temperature}: Quantitative, continuous data indicating the surrounding air temperature in Kelvin (K).
	\item \textbf{Time Stamp}: Qualitative, nominal data recording the date and time of each measurement.
\end{itemize}

The Month, Time of day and seasonality were extracted from the time stamp field to enhance the model's predictive accuracy. The dataset's temporal range spanned from 2020 to the present, offering a rich historical perspective on solar rad patterns under varying weather conditions. The dataset is summarized in the following Table \ref{tab:dataresumen}:

\begin{table}[ht]
	\centering
	\begin{tabular}{p{1cm}p{1.5cm} p{1.5cm}p{3cm}p{2.5cm}}
		\toprule
		{\textbf{Desc}} &  \makebox[1.5cm]{\centering \textbf{Month}} & \makebox[1.5cm]{\centering \textbf{Hour}} &  \makebox[3cm]{\centering \textbf{Irradiance (W/m²)}}   &  \textbf{Temperature K} \\
		\midrule
		count &  55057 & 55057 &   55057 &     55057 \\
		mean  &  6.07  	  & 11.53 &     141.93 &       299.77 \\
		std   &    3.23	  &	6.94 &     233.49 &         4.02 \\
		min   &    1 	  & 0.00 &      -9.00 &       291.85 \\
		25\%   &    3  	  & 6.00 &      -1.00 &       296.75 \\
		50\%   &     6	  & 12.00 &       0.00 &       298.45 \\
		75\%   &     9	  & 18.00 &     222.00 &       302.15 \\
		max   &   12  	  &	23.00 &    1459.00 &       314.85 \\
		\bottomrule
	\end{tabular}
	\caption{Data Resume}
	\label{tab:dataresumen}
\end{table}

\subsection{Data Visualization}

\begin{figure}[!ht]
	\includegraphics[width=8cm]{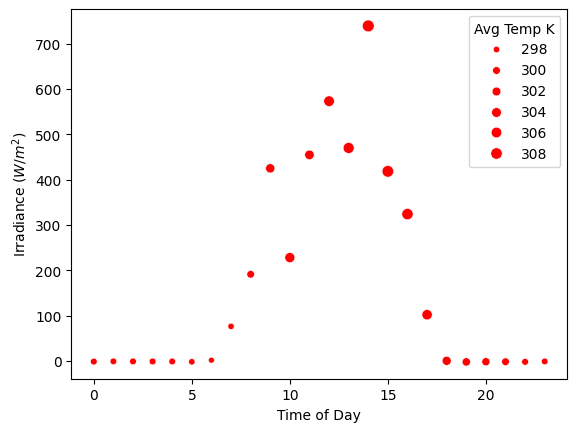}
	\centering
	\caption{Actual Solar Radiation on May 2, 2020}
	\label{fig:onedayradiation}
\end{figure}

The Figure \ref{fig:onedayradiation} illustrates the solar radiation measured throughout the day on May 2, 2020. The horizontal axis denotes the time of the day, and the vertical axis represents irradiance in $W/m^2$. Data points color and size depicting the average temperature in Kelvin per hour. Shades and sizes progress from light and small for cooler temperatures to dark and large for warmer temperatures, respectively, providing insights into the diurnal variation of solar irradiance and ambient temperature.

\begin{figure}[!ht]
	\includegraphics[width=8cm]{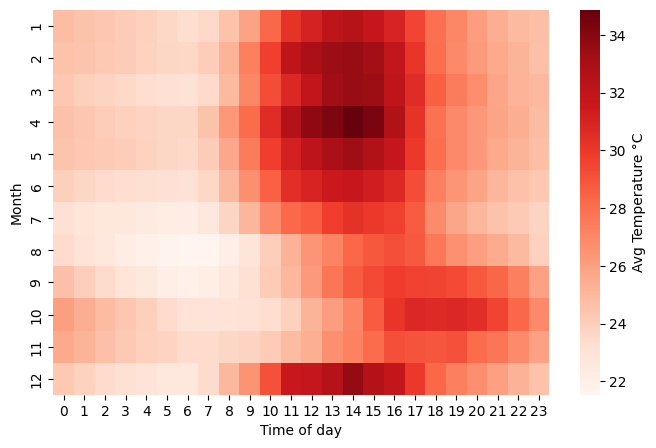}
	\centering
	\caption{Actual Solar Radiation on May 2, 2020}
	\label{fig:temp_hour_month}
\end{figure}

The Figure \ref{fig:temp_hour_month} represents the average hourly temperature for each month, as indicated by the color intensity. The horizontal axis lists the hours in a day, from 0 to 23, and the vertical axis enumerates the months. The gradation of colors, from light to dark, denotes the increase in temperature, measured in degrees Celsius. Notably, the map reveals that temperatures at midday during winter months are comparably high to day and night temperatures in other seasons, offering a clear depiction of temporal thermal patterns.

\subsection{Variables Selection}
\begin{figure}[ht]
	\includegraphics[width=7cm, height=6cm]{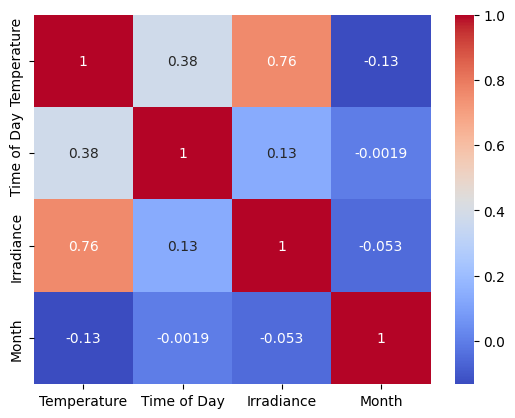}
	\centering
	\caption{Heatmap of Correlations between solar irradiance and environmental factors}
	\label{fig:correlation}
\end{figure}

The correlation analysis conducted on the dataset reveals several key insights into the relationships between solar irradiance and other measured variables. The heatmap of the correlation matrix, as depicted in Figure \ref{fig:correlation}, illustrates these relationships comprehensively.

The highest positive correlation observed with Irradiance is with Temperature, which has a coefficient of 0.76. This strong correlation suggests that as the temperature increases, there is a concomitant and substantial increase in Irradiance. Hence, Temperature is selected as a primary variable for predicting Solar Radiation due to its strong linear relationship.
Conversely, the variable Time of Day has a relatively lower positive correlation with Irradiance, indicated by a coefficient of 0.13. Although this positive correlation is present, it is considerably weaker, suggesting that the time of day, while relevant, is not as strongly predictive of Solar Radiation levels as Temperature is. This analysis forms the basis for selecting the most influential variables for the predictive model. The strong correlation of temperature and time of day with solar radiation underlines their importance in forecasting solar radiation levels at UTEQ. 

\subsection{Machine Learning Algorithms}
With the aim of choose the best algorithm to build our predictive model, a comprehensive comparative analysis was conducted using various machine learning algorithms. The algorithms included were: Linear Regression, a basic predictive analysis algorithm focusing on the linear relationship between dependent and independent variables \cite{montgomery2021linear}; Polynomial Regression, an extension of linear regression, which models a non-linear relationship between the dependent and independent variables \cite{seber2012linear}; K-Nearest Neighbors (KNN), a non-parametric method used for regression and classification tasks \cite{altman1992introduction}; Decision Tree Regressor, which models decisions and their possible consequences as a tree-like structure \cite{breiman1984classification}; Support Vector Regression (SVR) with both Linear and Radial Basis Function (RBF) kernels, a powerful technique derived from Support Vector Machines (SVM) for regression tasks \cite{drucker1997support}; Random Forest Regressor, an ensemble learning method based on decision tree algorithms \cite{breiman2001random}; and Gradient Boosting Regressor, an ensemble technique that builds models sequentially to minimize the loss function \cite{friedman2001greedy}. These models were evaluated to determine the most effective approach for predicting solar radiation levels at UTEQ.

\subsection{Configuration of the Machine Learning Algorithms}
In our analysis, we employed various machine learning algorithms, each with specific parameters, either set explicitly or left as default. The parameters for each algorithm are as follows:

\begin{itemize}
	\item \textbf{Linear Regression}: The \texttt{LinearRegression} algorithm was used with its default parameters, which include ordinary least squares fitting.
	
	\item \textbf{Polynomial Regression}: \texttt{PolynomialFeatures} with a degree of 2 was applied, enhancing the algorithm to capture non-linear relationships by raising input features to the specified degree.
	
	\item \textbf{K-Nearest Neighbors}: The \texttt{KNeighborsRegressor} was configured with \texttt{algorithm=`auto'}, which automatically chooses the appropriate algorithm based on the input data, \texttt{metric=`euclidean'} for Euclidean distance measurement, \texttt{weights=`distance'} to weight points by the inverse of their distance, and \texttt{n\_neighbors=10} to consider the 10 nearest neighbors.
	
	\item \textbf{Decision Tree Regressor}: This algorithm used \texttt{DecisionTreeRegressor} with \texttt{max\_depth=3}, limiting the depth of the tree to 3, and \texttt{random\_state=42} for reproducibility of results.
	
	\item \textbf{SVR Kernel Lineal}: The \texttt{SVR} algorithm with a \texttt{kernel=`linear'} was used, which applies a linear kernel function in Support Vector Regression.
	
	\item \textbf{SVR Kernel RBF}: Here, \texttt{SVR} was used with \texttt{kernel=`poly'}, employing a polynomial kernel function.
	
	\item \textbf{Random Forest Regressor}: The \texttt{RandomForestRegressor} was applied with \texttt{n\_estimators=100}, setting the number of trees in the forest to 100, and \texttt{random\_state=42} for consistent results across runs.
	
	\item \textbf{Gradient Boosting Regressor}: For this algorithm, \texttt{GradientBoostingRegressor} was used with \texttt{n\_estimators=100} to define the number of boosting stages, \texttt{learning\_rate=0.2} to control the contribution of each tree, \texttt{random\_state=42} for reproducibility, and \texttt{max\_depth=5} to set the maximum depth of the individual regression estimators.
\end{itemize}

\subsection{Evaluation Metrics}
To assess the performance of the built machine learning models in predicting solar radiation levels, the following evaluation metrics were applied:

\begin{itemize}
	\item \textbf{Mean Squared Error (MSE)}: It measures the average of the squares of the errors or deviations, that is, the difference between the estimator and what is estimated \cite{willmott2005advantages}. The MSE is calculated as:
	\begin{equation}
		MSE = \frac{1}{n}\sum_{i=1}^{n}(Y_i - \hat{Y}_i)^2
	\end{equation}
	where \( Y_i \) is the observed value and \( \hat{Y}_i \) is the predicted value.
	
	\item \textbf{Root Mean Squared Error (RMSE)}: It is the square root of the mean square error. The RMSE is a measure of the differences between values predicted by a model and the values observed \cite{chai2014root}. It is given by:
	\begin{equation}
		RMSE = \sqrt{MSE}
	\end{equation}
		
	\item \textbf{Mean Absolute Error (MAE)}: This metric summarizes the average magnitude of the errors in a set of predictions, without considering their direction \cite{willmott2005advantages}. It's calculated as:
	\begin{equation}
		MAE = \frac{1}{n}\sum_{i=1}^{n}|Y_i - \hat{Y}_i|
	\end{equation}
		
	\item \textbf{Coefficient of Determination (\( R^2 \))}: This metric provides an indication of the goodness of fit of a set of predictions to the actual values. In other words, it indicates how well unseen samples are likely to be predicted by the model  \cite{james2013introduction}. The \( R^2 \) score is calculated as:
	\begin{equation}
		R^2 = 1 - \frac{\sum_{i=1}^{n}(Y_i - \hat{Y}_i)^2}{\sum_{i=1}^{n}(Y_i - \bar{Y})^2}
	\end{equation}
	where \( \bar{Y} \) is the mean of the observed data.
\end{itemize}

These metrics are essential for evaluating and comparing the accuracy and effectiveness of different predictive models.

\section{Results}

\subsection{Machine Learning Models Evaluation}

The Table \ref{tab:ml_model_comparison} shows the results of the comparative analysis of machine learning models for predicting solar radiation, demonstrate a range of performances as measured by the Mean Squared Error (MSE), Root Mean Squared Error (RMSE), Mean Absolute Error (MAE), and the Coefficient of Determination ($R^2$).

\begin{table}[ht]
	\centering
	\begin{tabularx}{\textwidth}{lXXXX}
		\toprule
		\textbf{Model}					   & \textbf{MSE}   & \textbf{RMSE} & \textbf{MAE}  & \textbf{$R^2$}   \\
		\midrule
		Linear Regression                  & 21773.10       & 147.56        & 105.10        & 0.60          \\ 
		Polynomial Regression              & 16268.56       & 127.55        & 82.69         & 0.70          \\ K-Nearest Neighbors                & 14625.13       & 120.93        & 59.33         & 0.73          \\ Decision Tree Regressor            & 17540.69       & 132.44        & 78.75         & 0.68          \\ SVR Kernel Lineal                  & 23889.29       & 154.56        & 101.22        & 0.56          \\ SVR Kernel RBF                     & 29976.37       & 173.14        & 104.11        & 0.44          \\ Random Forest Regressor            & 14106.10       & 118.77        & 58.56         & 0.74          \\ \textbf{Gradient Boosting Regressor}        &\textbf{ 13112.21}       & \textbf{114.51}        & \textbf{56.87}         & \textbf{0.72}          \\ 
		\bottomrule
	\end{tabularx}
	\caption{Results of the Machine Learning Models evaluation}
	\label{tab:ml_model_comparison}
\end{table}

\begin{figure}[ht]
	\centering
	\includegraphics[width=10cm, height=7cm]{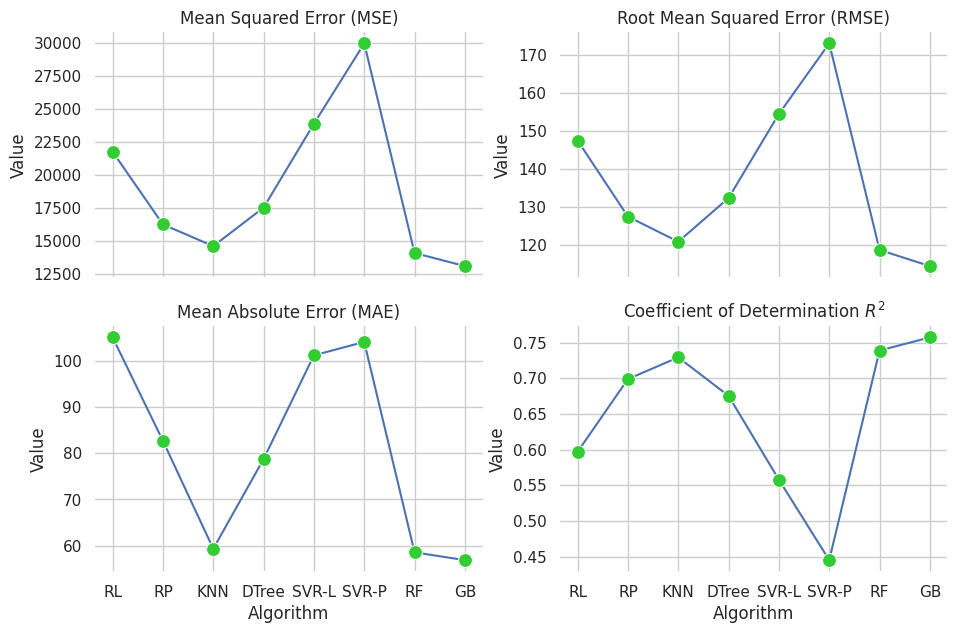}
	\caption{Results of the Machine Learning Models evaluation}
	\label{fig:metricas_comparison}
\end{figure}

In the Figure \ref{fig:metricas_comparison} the performance of each model is marked by a green dot, allowing for a visual comparison of their predictive accuracy and fit.

The Linear Regression model exhibits a moderate level of accuracy with an $R^2$ of 0.60, suggesting that approximately 60\% of the variance in the radiation data is predictable from the model. However, its higher MSE and RMSE values indicate a greater average error in the predictions. Polynomial Regression shows an improvement over Linear Regression, with a lower MSE and a higher $R^2$ of 0.70, which implies a better fit to the data and the model's ability to capture more complex relationships. K-Nearest Neighbors (KNN) achieves an even lower MSE and the highest $R^2$ value among the first three models, reflecting its capability to closely predict the actual values. The localized nature of KNN likely contributes to its relatively high accuracy. Decision Tree Regressor offers a balance between complexity and interpretability, resulting in a decent $R^2$ of 0.68, but not outperforming KNN in terms of error metrics. SVR with a Linear Kernel and SVR with a Polynomial Kernel (RBF) struggle with this dataset, as indicated by their higher MSE and lower $R^2$ values, especially the SVR with RBF kernel, which has the highest errors and lowest $R^2$ value of 0.44, suggesting a poor fit. Random Forest and Gradient Boosting Regressors show the best performance among all the models, with the lowest MSE and RMSE values and the highest $R^2$ values of 0.74 and 0.72, respectively. These ensemble methods, known for their high accuracy, confirm their effectiveness in the task of solar radiation prediction.

\subsection{Performance of the Solar Radiation Prediction}

\begin{figure}[ht]
	\includegraphics[width=\textwidth, height=12cm]{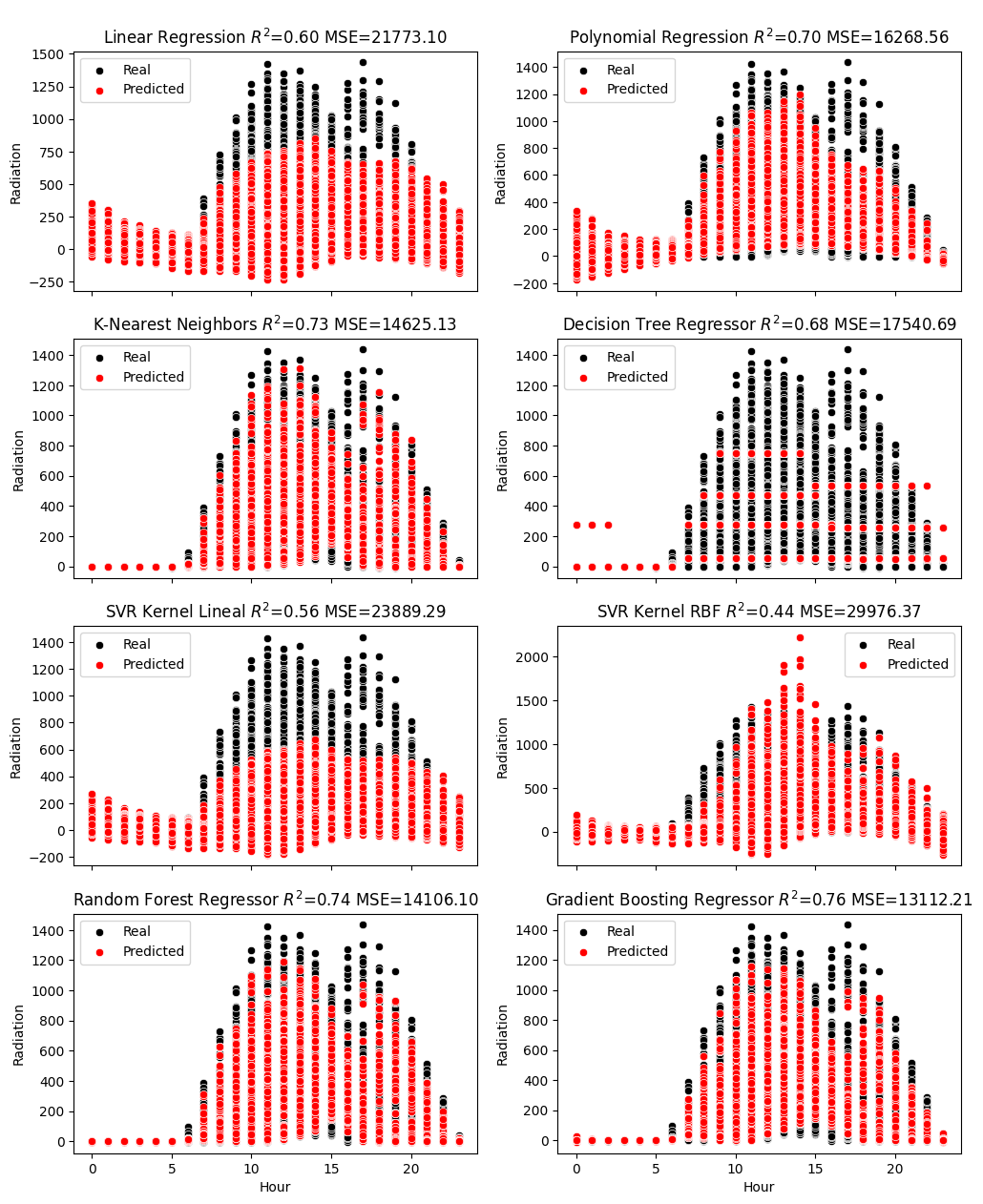}
	\centering
	\caption{Scatter plots comparing the real and predicted solar radiation values for different machine learning models}
	\label{fig:comparative}
\end{figure}

The comparative analysis of machine learning models for predicting solar radiation reveals varying degrees of accuracy and predictive performance. As illustrated in the scatter plots (see Figure \ref{fig:comparative}), Linear Regression and SVR with Linear Kernel show a broader dispersion of predicted values from the actual measurements, resulting in lower \(R^2\) values of 0.60 and 0.56, respectively, and higher MSE, indicating a less precise fit to the data. Polynomial Regression and Decision Tree Regressor improve upon these metrics with \(R^2\) values of 0.70 and 0.68, respectively, suggesting a better but still not optimal fit. The K-Nearest Neighbors model exhibits a tighter cluster of predictions around the actual values with an \(R^2\) of 0.73, indicating a good fit to the data. However, the Random Forest and Gradient Boosting models outperform the others with the highest \(R^2\) values of 0.74 and 0.76, respectively. These models not only have a tighter clustering of predictions but also the lowest MSE, indicating a superior predictive performance.

\begin{figure}[ht]
	\centering
	\includegraphics[width=\textwidth, height=8cm]{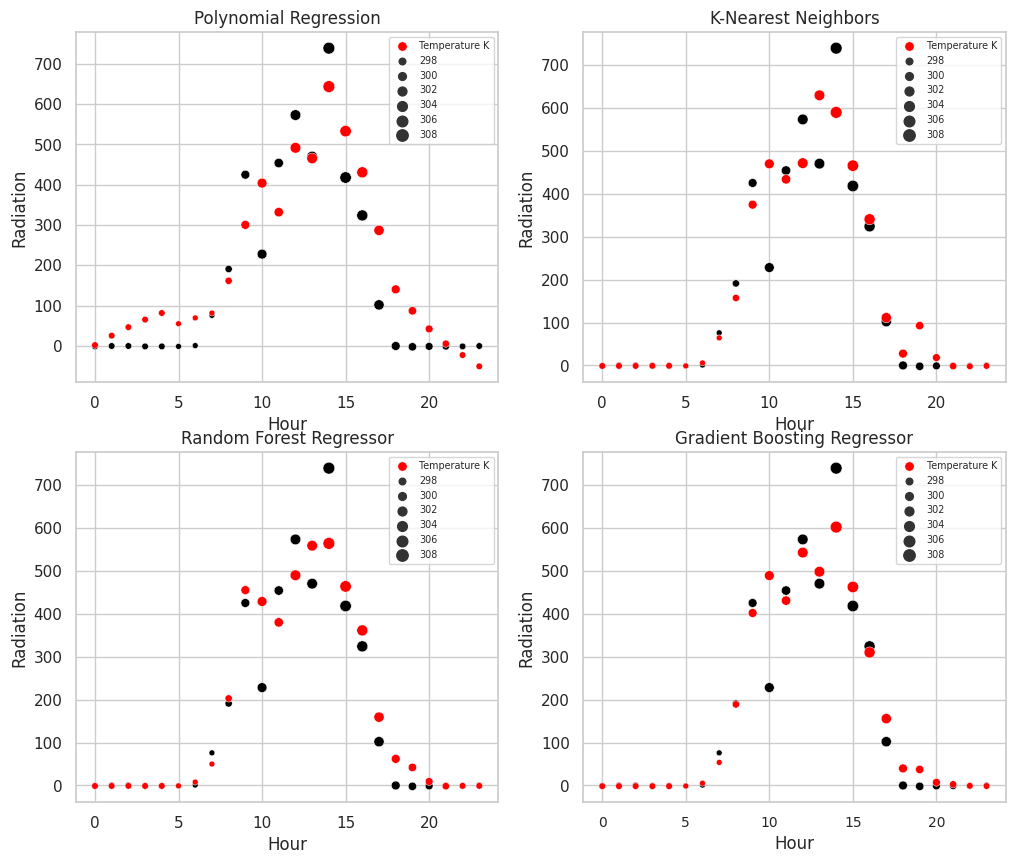}
	\caption{Scatter plots comparing actual and predicted solar radiation levels using four machine learning algorithms. The actual data is represented in black, while the predictions are in red, with varying dot sizes corresponding to different temperature levels.}
	\label{fig:solar_radiation_prediction}
\end{figure}

The scatter plots in Figure \ref{fig:solar_radiation_prediction} provide a visual comparison between actual and predicted solar radiation levels for a specific day, May 2nd, 2020. 

The Polynomial Regression, K-Nearest Neighbors (KNN), Random Forest Regressor, and Gradient Boosting Regressor models were evaluated. The actual measurements are marked in black and the predictions in red, with the dot sizes indicating varying temperatures. It is observed that the Random Forest and Gradient Boosting models closely align with the actual data, especially during peak radiation hours, suggesting their higher efficacy in capturing the non-linear patterns associated with solar radiation. The slight discrepancies noted in the Polynomial Regression and KNN predictions during peak hours warrant further investigation and potential model refinement for improved accuracy.

\subsection{Interactive tool for Solar Radiation Forecasting}
With the aim of provide an easy tool to assess the performance of the built Machine Learning models, we have developed an interactive web application for solar radiation forecasting, available at \url{https://solarradiationforecastinguteq.streamlit.app/}, offers real-time predictions based on temperature forecasts from the Meteosource API (\url{https://www.meteosource.com/}). It features a user-friendly interface that allows the selection of the best-performing models as identified in our comparative analysis: Polynomial Regression, K-Nearest Neighbors, Random Forest Regressor, and Gradient Boosting Regressor.

\begin{figure}[ht]
	\centering
	\includegraphics[width=\textwidth, height=8cm]{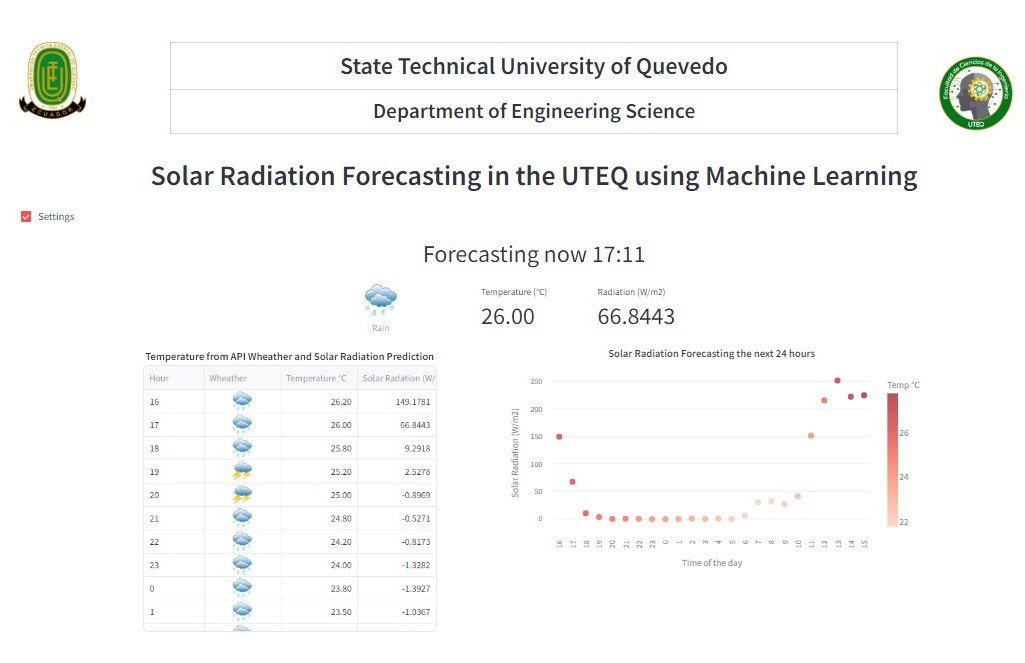}
	\caption{A web application to real-time solar radiation forecasting based on machine learning models.}
	\label{fig:webprediction}
\end{figure}

The Figure \ref{fig:webprediction} shows the main dashboard, it provides a current forecast snapshot, displaying the temperature and predicted solar radiation at a specific time, alongside a graphical representation of the 24-hour forecast. The plots illustrate the predictions against the actual solar radiation data, with the models' performance visible through the close alignment of predicted (in red) and actual (in black) values. The integration of temperature data as dot sizes in the scatter plots offers an intuitive understanding of the relationship between temperature and solar radiation levels throughout the day.

\section{Conclusions and Future Works}
This study demonstrated the viability of using machine learning models for predicting solar radiation at UTEQ. Among the models tested, Gradient Boosting Regressor exhibited superior performance, closely followed by the Random Forest Regressor. These models effectively captured the non-linear patterns in solar radiation, as evidenced by their low MSE and high R² values. The integration of these models into a user-friendly web application underscores their practical utility in real-time solar radiation forecasting, aiding in efficient solar energy management.

Future research should focus on enhancing the accuracy of the predictive models further. This could involve exploring more sophisticated machine learning algorithms or deep learning techniques. Additionally, integrating a larger dataset with more diverse environmental variables could provide a more comprehensive understanding of the factors influencing solar radiation. Another promising avenue is the application of these models in other geographical locations to validate their generalizability and effectiveness. Finally, real-world implementation and feedback will be crucial in refining the models and maximizing their utility in practical applications.

\bibliographystyle{splncs03_unsrt} 
\bibliography{refs}

\end{document}